\documentclass[10pt]{article}
\usepackage[letterpaper]{geometry}
\usepackage{hicss}
\usepackage{times}
\usepackage[none]{hyphenat}
\usepackage{url}
\usepackage{latexsym}
\usepackage{minted}
\usepackage{indentfirst}
\usepackage{graphicx}
\graphicspath{{images/}}
\usepackage[
    style=apa,
  ]{biblatex}
\usepackage{graphicx}

\title{Distribution-Free Process Monitoring with Conformal Prediction}

\author{Christopher Burger \\
 The University of Mississippi \\
 {cburger@olemiss.edu} \\ \\
}
\date{}

\begin{document}
\maketitle
\begin{abstract}
Traditional Statistical Process Control (SPC) is essential for quality management but is limited by its reliance on often violated statistical assumptions, leading to unreliable monitoring in modern, complex manufacturing environments. This paper introduces a hybrid framework that enhances SPC by integrating the distribution free, model agnostic guarantees of Conformal Prediction. We propose two novel applications: Conformal-Enhanced Control Charts, which visualize process uncertainty and enable proactive signals like `uncertainty spikes', and Conformal-Enhanced Process Monitoring, which reframes multivariate control as a formal anomaly detection problem using an intuitive p-value chart. Our framework provides a more robust and statistically rigorous approach to quality control while maintaining the interpretability and ease of use of classic methods.
\end{abstract}

\subsubsection*{Keywords:}

Statistical Quality Control, Statistical Process Control, Conformal Prediction, Anomaly Detection, Quality Management

\section{Introduction}



The operational landscape of modern manufacturing and business is defined by an increasing degree of complexity, automation, and data generation. Production systems are often no longer simple, linear processes but intricate, multi stage networks that generate high dimensional data streams in real time. This presents a significant challenge for quality control, which has become a major business strategy for increasing productivity and gaining (or maintaining) a competitive advantage. The failure to maintain quality can lead to substantial economic losses from scrap and rework, warranty claims, and reputational damage and these problems are enumerated early on in most industrial and systems engineering curricula \parencite{turner1992introduction}. In safety critical industries, such as aerospace or medical devices, a single quality failure can have catastrophic consequences, like with UA Flight 232 \parencite{NTSB1990UA232} or the recent CV-22 Osprey Crash \parencite{AFSOC2024CV22,Reuters2024CV22}. With the increasing complexity in modern manufacturing there is now a need for quality control methodologies that are robust and reliable enough to manage difficulties presented in modern production environments. 

For nearly a century, Statistical Process Control (SPC) has been a cornerstone of quality management, providing an intuitive framework for monitoring production by distinguishing between natural process variation and problematic "special cause" variation that signals a plausible malfunction. Many conventional Statistical Process Control (SPC) charts are designed with the assumption that the process data is normally distributed. This is a foundational principle for classical parametric statistical methods, which are based on assumptions of normality, independence, and constant mean and variance. The very construction of the popular Shewhart control chart is based on the properties of the normal distribution, where control limits are typically set at three standard deviations from the mean, a practice that captures approximately 99.7\% of the data if the process is normal. In conventional SPC, the pattern of variation from "common causes" is often assumed to follow a normal distribution \parencite{turner1992introduction}.  

These these assumptions are significant limitations, as data from  industrial processes can be non-normal, skewed, or multimodal \parencite{smaj_nonpara,article}. Applying traditional control charts in these common scenarios can lead to unreliable control limits, resulting in an increased rate of false alarms or, more dangerously, a failure to detect true process shifts, thereby undermining the very purpose of the control system. While non-parametric alternatives exist, they often lack the statistical power of their parametric counterparts or are not easily adapted to complex, multivariate settings \parencite{smaj_nonpara,article}. 

This paper proposes a hybrid framework that enhances the proven, practical structure of SPC with the statistical rigor of Conformal Prediction (CP), a modern machine learning technique for uncertainty quantification. The central advantage of CP is its ability to produce prediction intervals with a mathematically guaranteed, user specified error rate while making fewer assumptions, and importantly, none are needed about the underlying data distribution \parencite{vovk2022algorithmic}. Furthermore, the CP framework is model agnostic, meaning it can be used in combination with any other predictive algorithm, from simple regression to advanced deep learning models \parencite{vovk2022algorithmic}. By replacing the limiting parametric assumptions of traditional SPC with the distribution free guarantees of CP, our method provides a more robust, reliable, and universally applicable approach to quality control that is fundamentally better suited for the complex data generated by modern industrial processes.  

Our primary contributions are the development and exploration of two novel applications of this integrated framework. First, we introduce Conformal-Enhanced Control Charts, which visualize process stability by plotting prediction intervals over time instead of single data points. This provides a richer, more intuitive display of process behavior and enables new types of control signals, such as `uncertainty spikes', which can serve as leading indicators of process instability. Second, we propose Conformal-Enhanced Process Monitoring, a method that reframes multivariate quality control as a formal anomaly detection problem. This approach uses the conformal p-value (defined in Section \ref{sec:background_cp}) as a unified, easily interpretable metric of overall process health. This can then be visualized on a simple p-value chart, making the monitoring of highly complex systems accessible to operators without specialized statistical knowledge. Together, these contributions establish a quality system that is more accurate, proactive, and statistically sound.

\section{Background and Related Work}

\subsection{Traditional Methods in Statistical Process Control}
The foundation of modern quality control is Statistical Process Control (SPC), a discipline pioneered by Walter A. Shewhart in the 1920s. SPC provides a statistical framework for monitoring and controlling a process to ensure it operates at its full potential to produce conforming products. Shewhart's core contribution was the distinction between two types of process variation: "common cause" variation, which is inherent and natural to a stable process, and "special cause" (or assignable) variation, which stems from external, identifiable sources indicating a process is out of statistical control. The practical implementation of SPC is a two step activity: Step one uses a historical dataset to understand the process and establish stable control limits, while step two then uses these limits for ongoing, real-time monitoring to detect any new special causes of variation. The classic reference for the discipline can be found in \textit{Introduction to Statistical Quality Control} \parencite{montgomery2019introduction}.

The most common tool in SPC is the control chart, which graphically displays a quality characteristic over time against these control limits. However, classical charts like the Shewhart $\bar{X}$ chart are parametric, relying on the (often incorrect) assumption that the process data is well approximated by a normal distribution. Using these charts with non-normal data can lead to unreliable control limits and erroneous conclusions. To address this, a variety of non-parametric (or distribution free) control charts were developed. These charts, such as Sign charts and Wilcoxon rank-sum charts, do not require distributional assumptions and instead often rely on ranks or signs of the data, providing more robust performance when process knowledge is limited. While valuable, these methods can be less powerful than their parametric counterparts or difficult to adapt to complex multivariate settings \parencite{dfmSPMC}.

\subsection{Conformal Prediction}\label{sec:background_cp}
Conformal Prediction (CP) is a modern machine learning framework that provides rigorous, distribution free uncertainty quantification. Developed by Vovk, Gammerman, and Shafer \parencite{vovk2022algorithmic,shafer2008tutorial}, CP transforms the point predictions of any underlying algorithm into prediction sets (for classification) or intervals (for regression). The key feature of these sets is a mathematically guaranteed marginal coverage rate; for a user specified error level $\alpha$, the true outcome will be contained within the prediction set with probability at least $1-\alpha$. This guarantee is model-agnostic and holds for any data distribution, requiring only the minimal assumption that the data is exchangeable, which can be viewed as a statistical formalization of a stable and in-control process.  

The mechanism of CP relies on a user defined nonconformity score, which measures how anomalous a data point is relative to a set of typical examples. A common choice for regression is the absolute residual between the predicted and true values. In the most practical variant, Inductive Conformal Prediction (ICP), a reference dataset is split into a proper training set and a calibration set. The model is trained on the training set, and nonconformity scores are then calculated for each example in the calibration set. The $(1-\alpha)$ quantile of these calibration scores determines the size of the prediction interval for a new test point. This split approach is computationally efficient, making it suitable for real time applications, unlike the original transductive formulation which required retraining for every new point \parencite{angelopoulos2022gentleintroductionconformalprediction, ndiaye2019computingconformalpredictionset}. 

An important component of the conformal prediction framework is the conformal p-value which measures how unusual a new data point is compared to a set of past observations, based on a nonconformity score (e.g., error or residual). It is calculated by comparing the new point’s score to those from a calibration set, and represents the proportion of calibration scores that are greater than or equal to it. Unlike the frequentist p-value, which represents the probability of obtaining a result as extreme as the observed one assuming a specific null hypothesis and distribution, the conformal p-value makes minimal assumptions (primarily that the data are exchangeable). This makes conformal p-values more intuitive for end users, as they can be interpreted simply as `how typical or atypical' a new observation is compared to past data, without needing to invoke abstract null hypotheses or reference distributions. This provides a substantial benefit, as frequentist p-values are notoriously difficult for both non-statistician researchers \parencite{wasserstein2016asa} and surprisingly even statisticians themselves \parencite{mcshane2017statistical} to use correctly.

While the direct application of CP to enhance traditional SPC charts is a novel contribution of this work, the underlying techniques have been successfully applied in related industrial and time-series contexts. Conformal methods are used for anomaly detection in large scale systems and for detecting abnormal trajectories in surveillance data \parencite{phdthesis}. In manufacturing, CP has been proposed for reliable surface defect detection, where it provides statistically rigorous guarantees on error rates \parencite{shen2025conformalsegmentationindustrialsurface}. Other applications include commercial farming \parencite{melki2025precision}, semiconductor manufacturing \parencite{akpabio2022ler}, and industrial recycling \parencite{dos2024scrap}. These applications demonstrate that despite the relatively recent adoption of conformal methods, their practicality has shown value to industry.

\section{Conformal Enhanced Control Charts}
The principles of Statistical Process Control and Conformal Prediction, while developed in different domains, are complementary. By integrating them, it is possible to create a new process monitoring tools that are more statistically rigorous, and so better suited for the complexities of modern manufacturing. This section proposes the concept of a Conformal-Enhanced Control Chart, detailing its construction and implementation. Our approach maintains the familiar visual representation of traditional control charts while replacing parametric assumptions with distribution-free conformal prediction methods for determining control limits.
\subsection{Core Framework}
The construction of a conformal-enhanced chart involves three main steps for establishing the control limits, followed by a final step for real time monitoring and visualization.
\subsubsection*{Initialization:}

The process begins by collecting a sequence of calibration observations. Calibration observations are historical data points collected when the process was known to be operating in a stable, in-control state. These serve as the baseline that defines normal process behavior. Let $X_1, X_2, ... X_n$ be this sequence of calibration observations from an in-control process. 





\begin{figure*}[htbp]
  
    \centering
        \includegraphics[height=0.46\textheight, width=0.9\textwidth]{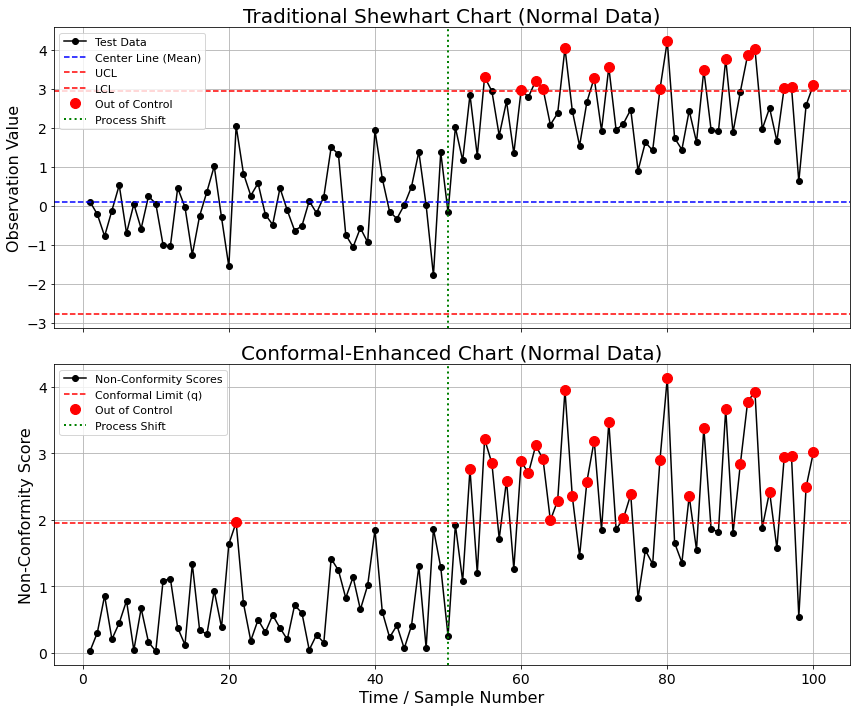}  

          \caption{Comparison between a traditional chart and a conformally enhanced chart for normal data}
    \label{fig:norm}
\end{figure*}

\subsubsection*{Non-conformity Measure Selection:}
Next, a non-conformity score (NCS) function, $s(X)$, is defined. This function's purpose is to quantify how unusual or atypical a new observation is compared to the baseline established by the calibration set. A higher score indicates a greater deviation from typical behavior. If a new observation has a high non-conformity score compared to calibration data, this is cause to investigate further.
\\

\noindent We then choose a non-conformity measure that is dependent on our goal. \begin{itemize}
    \item For individual observations: \\ \\ $s(X_i) = |X_i - median(X_1, ... X_n)|$
    \item For subgroups: \\ \\ $s(\bar{X}_i) = |\bar{X}_i - median(\bar{X}_1, ... \bar{X}_n)|$
    \item For variability: \\ \\ $s(R_i) = |R_i - median(R_1, ... R_n)|$ (Where $R$ is the range)
\end{itemize}

We have chosen the median here as it is robust to `incongruities' within the data, like extreme outliers or severe departures from normality. As our goal is to reduce the number of assumptions and be able to deal with non-normal data, this statistic is a natural choice, though any measure of central tendency could be substituted.

\subsubsection*{Determining the Control Limits:}

With the NCS defined, the control limit is determined from the calibration data.
\begin{enumerate}
    \item Compute the non-conformity scores for all observations in the calibration set: $S = \{s(X_1), s(X_2),..., s(X_n)\}$. 
    \item Choose a desired false alarm rate, $\alpha$. This value represents the acceptable long run probability of a Type 1 error (a false alarm). For example, to emulate the  $\pm3\sigma$ limits of a traditional Shewhart chart, which cover approximately 99.73\% of in-control data, one would select $\alpha = 0.0027$. 
    \item The upper conformal threshold, $q$, is set as the $(1-\alpha)$ quantile of the calculated scores in $S$. More precisely, $q$ is the $\lceil(1-\alpha)(n+1)\rceil$-th smallest value in the sorted list of scores. The value $q$ now serves as the single, statistically rigorous control limit.
\end{enumerate}

\begin{figure*}[htbp]
  
    \centering
        \includegraphics[height=0.46\textheight, width=0.9\textwidth]{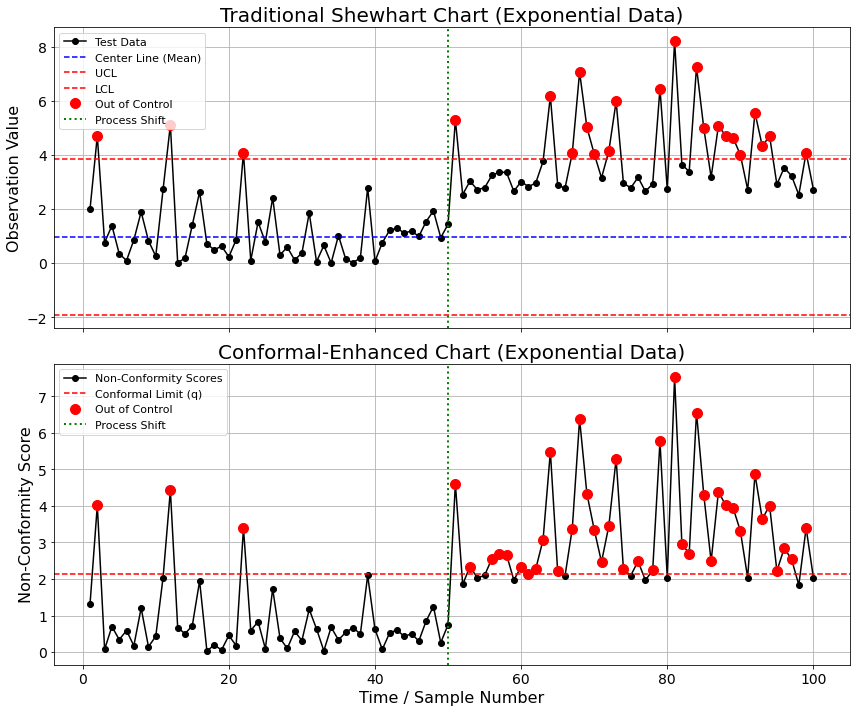}  

          \caption{Comparison between a traditional chart and a conformally enhanced chart for exponential data}
    \label{fig:exp}
\end{figure*}

\subsection{Visualization}
With the control limits calculated, we have all we need to construct a conformal enhanced control chart. We can then use the chart for prospective monitoring. For any new observation $X_{new}$, its non-conformity score $s(X_{new})$ is computed and if $s(X_{new}) > q$, the process is flagged as being out of control. This provides a direct, distribution free alternative to a traditional Shewhart chart. We can see in Figures \ref{fig:norm} and \ref{fig:exp} a comparison between a standard Shewhart chart and our conformally modified version. Here we simulate a shift in the mean of the generating distribution and compare the number of points flagged as out of control. Our conformally enhanced chart shows greater sensitivity to true shifts over the standard formulation. Note that Figure \ref{fig:norm} is still normally distributed data, just because the data \textit{does} follow the standard assumptions does not inherently reduce the effectiveness of our method.

\begin{figure*}[htbp]
   
    \centering
        \includegraphics[height=0.46\textheight, width=0.9\textwidth]{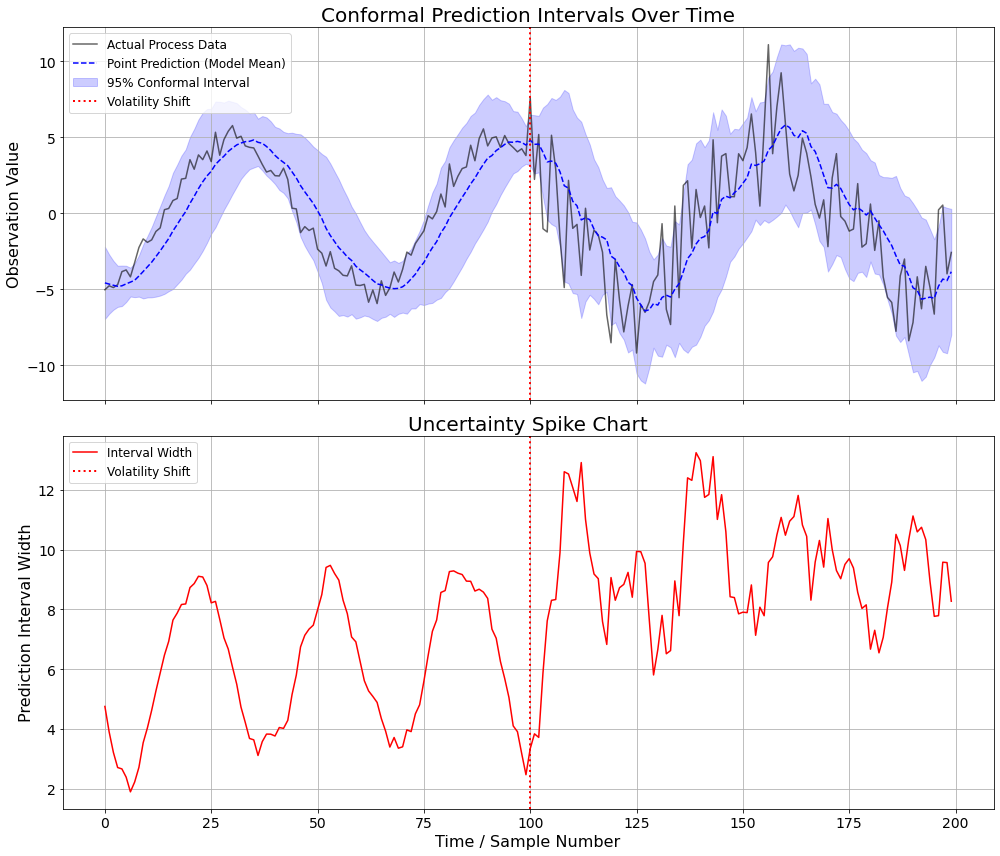}  

         \caption{Conformal interval chart and Uncertainty spike chart}
    \label{fig:usc}
\end{figure*}

However, our framework allows for further improvements to visualization and interpretation. Instead of plotting a single point, a Conformal-Enhanced Control Chart can plot a conformal prediction interval for each new observation. This transforms the chart from a simple line of points into a dynamic band or ribbon whose width represents quantified uncertainty (Figure \ref{fig:usc}). This enhanced visualization lets us decouple the signals related to process shifts from signals related to process uncertainty. While a shift in the location of the interval still signals a change in the process mean, a new signal emerges when using model based non-conformity scores.

If a predictive model $\hat{y}(x)$ is available (e.g., predicting a quality characteristic $y$ from process parameter(s) $x$), the NCS can be defined as the model's prediction error, such as the absolute residual $s(x,y) = |y - \hat{y}(x)|$. The resulting conformal interval, $[\hat{y}(x_{new}) - q, \hat{y}(x_{new}) + q]$, will now be centered on the model's real time prediction.

By using a normalized non-conformity score that adapts to local conditions (e.g., $s(x,y) = \frac{|y - \hat{y}(x)|}{\hat{\sigma}(x)}$, where $\hat{\sigma}(x)$ is a model's estimate of local variability), the width of the prediction interval becomes adaptive. A sudden increase in the interval's width (an `uncertainty spike') can be viewed as a new type of control signal (Figure \ref{fig:usc}). This signal indicates that the model is encountering unfamiliar or noisy process conditions, even if the quality characteristic itself has not yet deviated from its target. This serves as a  leading indicator of potential instability, allowing for proactive investigation and corrective action before a process shift results in out of spec product. This then transforms process control from a reactive to a proactive discipline.

\section{Conformal-Enhanced Process Monitoring}
While the Conformal-Enhanced Control Chart provides a powerful, distribution free method for monitoring a single quality characteristic, modern manufacturing processes are rarely defined by a single variable. Instead, their state is described by the complex and correlated behavior of numerous process parameters captured by sensors in real time. Attempting to monitor each of these variables with an individual chart is not only impractical but also statistically unsound, as it fails to account for their interdependencies and can lead to a high rate of false alarms.This section extends the conformal framework to address this by reinterpreting process monitoring as a formal, statistically rigorous anomaly detection problem.

An important consideration is that the state of a complex process at any given moment is often best described by a high dimensional vector of sensor readings, including temperatures, pressures, flow rates, and chemical concentrations, etc. These variables are rarely independent. A change in one parameter can have cascading effects on others. Traditional Multivariate SPC (MSPC) methods, like the Hotelling $T^2$ chart, were developed to address this by monitoring variables jointly. However, these methods typically suffer from their own restrictive assumptions, such as the requirement for data to follow a multivariate normal distribution, and their outputs can be difficult for operators to interpret, since a single $T^2$ statistic does not indicate which variable is responsible for an out-of-control signal.  

\begin{figure*}[htbp]
    
    \centering
        \includegraphics[height=0.37\textheight, width=0.9\textwidth]{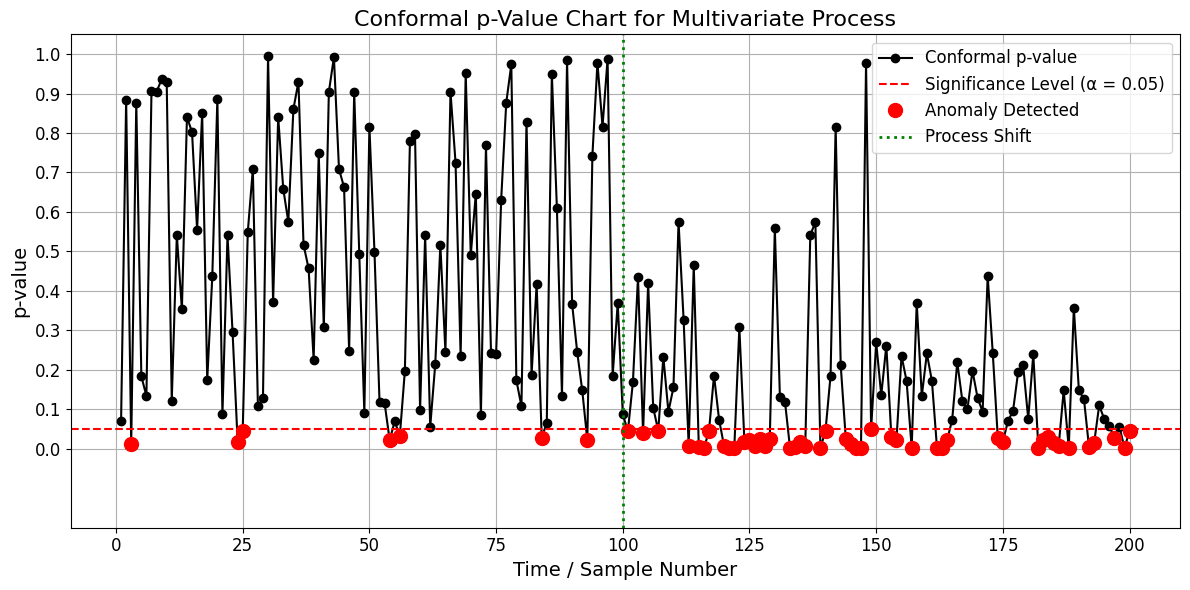}  
        \caption{Conformal p-value chart}
    \label{fig:cpc}
\end{figure*}

\subsection{Process Monitoring as Conformal Anomaly Detection}

A more powerful and flexible approach is to reframe the entire problem of process monitoring as one of online anomaly detection. In this paradigm, the goal is not to track individual variables but to answer a single, generalized question for each new snapshot of the process: is this state `normal' (in-control) or `anomalous' (out of control)? Conformal prediction provides a framework to answer this question. Our methodology mirrors the two step SPC structure:

\subsubsection*{Training and Calibration:} 

We first collect a reference dataset consisting of many (possibly) high dimensional vectors, where each vector $\mathbf{x}_i$ represents a complete snapshot of all monitored process variables at a time when the process was known to be operating in a stable, in-control state.

Then an unsupervised anomaly detection model is trained on a portion of this reference data. Unlike supervised models, these algorithms do not require labeled examples of `failures'. Instead, they learn the underlying structure and patterns of the `normal' data. We do not focus on any particular model here, the choice of which will be determined by concerns that may differ between applications. But a wide range of powerful models can be used for this task, from (comparatively) simple support vector machines to autoencoders.

We then define a nonconformity score based on the output of the unsupervised model. Regardless of the model choice, a higher score signifies a greater degree of anomaly.
Finally, we perform calibration where a separate calibration set of in-control process vectors is passed through the trained anomaly detection model. The nonconformity score is calculated for each vector, creating an empirical distribution of scores that correspond to normal operating conditions.

\subsubsection*{Monitoring and Visualization:} As each new process state vector $\mathbf{x}_t$ is captured at time $t$, it is fed into the trained model to calculate its nonconformity score, $s_t$. This new score $s_t$ is compared against the distribution of scores from the calibration set to compute its conformal p-value. This p-value represents the statistical evidence for the new process state being an outlier relative to historical normal operations.

The output of this conformal anomaly detection process—a time series of p-values—can be visualized in a new and exceptionally powerful type of control chart: the Conformal p-Value Chart (Figure \ref{fig:cpc}. These charts are simple in construction and in interpretation. The y-axis represents the p-value, ranging from 0 to 1. The x-axis represents time or sample number. There is only a single control limit: a horizontal line drawn at the user specified significance level, $\alpha$ (e.g., $\alpha = 0.05$ or $\alpha = 0.01$).

The interpretation is direct and unambiguous. The process is considered in-control as long as the plotted p-values remain above the $\alpha$ line. Any p-value that dips below this line is a statistically valid signal that the current process state is anomalous with a long-run false alarm rate controlled at $\alpha$. The actual value of the p-value provides a continuous measure of evidence; a p-value of 0.001 is a much stronger signal of an anomaly than a p-value of 0.04.

This approach offers a advantage in operational settings. The challenge with traditional MSPC tools like the Hotelling $T^2$ chart is their inherent complexity. Interpreting the chart requires specialized statistical knowledge, and diagnosing the root cause of a signal is a nontrivial subsequent step. The conformal anomaly detection framework creates a powerful abstraction layer. All the complexity of the high dimensional data and the sophisticated underlying machine learning model is encapsulated into a single metric: the conformal p-value.

A process engineer or a factory floor operator does not need to understand the architecture of a variational autoencoder or the mathematics of a support vector machine. They only need to be trained on a single, simple rule: `If the point on this chart drops below the red line, the system is signaling a process anomaly with a predefined and guaranteed error rate.' This makes the monitoring of complex, multivariate processes as intuitive as reading a basic Shewhart chart. This allows for deployment state of the art, distribution free MSPC at scale within an organization as it separates the task of model development (generally performed by those with statistical training) from the task of real time monitoring (generally performed by process experts).


\section{Conclusion}
The integration of Statistical Process Control and Conformal Prediction offers a significant advancement in quality management. By augmenting the time tested, practical framework of SPC with the rigorous, distribution free guarantees of modern machine learning, we have proposed a methodology that is more reliable, informative, and adaptive to the complexities of contemporary industrial processes. Our approach preserves the intuitive, visual nature of control charts while replacing their rigid parametric assumptions. The introduction of Conformal-Enhanced Control Charts and Conformal-Enhanced Process Monitoring provides practitioners with powerful new tools to not only detect process shifts but also to proactively manage uncertainty and monitor complex, high dimensional systems in an easily interpretable manner.

\subsection{Limitations and Future Work}
Despite its advantages, the proposed framework has limitations that offer avenues for future research. The validity of any conformal method is  dependent on the quality and representativeness of the initial calibration dataset. If the initial data is not truly reflective of an in-control process, the statistical guarantees will be compromised. Additionally, while our proposed non-conformity scores are simple and robust, their ability to produce tight, informative prediction intervals is heuristic. The optimization of these scores for specific industrial applications remains an important area for exploration, likely requiring a combination of statistical expertise and domain knowledge to produce substantive benefits. Future work should focus on the next logical step, the development of more sophisticated, adaptive non-conformity scores that can learn from process data to improve efficiency. Additionally, the use and subsequent validation of these ideas in real manufacturing settings is essential to demonstrate their practical application.
\\

\noindent \textbf{Data Availability Statement}: All data and materials used in this study are openly available in a public repository: {\small \url{https://github.com/christopherburger/ConformalSPC}}
\\

\noindent \textbf{Acknowledgments}: The authors have no acknowledgments to declare.


\printbibliography

\end{document}